\documentclass{article}

\usepackage[final]{neurips_2020}

\usepackage[utf8]{inputenc} 
\usepackage[T1]{fontenc}    
\usepackage{url}            
\usepackage{booktabs}       
\usepackage{amsfonts}       
\usepackage{nicefrac}       
\usepackage{microtype}      
\usepackage{amsmath,amssymb}
\usepackage{mathtools}
\usepackage{amsthm}
\usepackage{bm}
\usepackage{xcolor}
\usepackage[english]{babel}
\usepackage{graphicx}
\usepackage{caption}
\usepackage{subcaption}
\usepackage{wrapfig}
\usepackage{makecell}
\usepackage{multirow}
\usepackage[misc]{ifsym}
\usepackage[colorlinks,linkcolor=red]{hyperref}

\theoremstyle{definition}

\theoremstyle{definition}

\theoremstyle{definition}

\theoremstyle{definition}

\makeatletter
\def\mathcolor#1#{\@mathcolor{#1}}
\def\@mathcolor#1#2#3{%
  \protect\leavevmode
  \begingroup
    \color#1{#2}#3%
  \endgroup
}
\makeatother

\title{Towards Good Practices of U-Net for\\ Traffic Forecasting}

\author{%
 Jingwei Xu$^1$,
 Jianjin Zhang$^2$\thanks{Correspondence to: jianjzh@microsoft.com} ,
 Zhiyu Yao$^3$,
 Yunbo Wang$^1$\\
 $^1$Shanghai Jiao Tong University
 $^2$Microsoft Corporation
 $^3$Tsinghua University
}

\begin{document}

\maketitle

\begin{abstract}
This technical report presents a solution for the 2020 Traffic4Cast Challenge.
We consider the traffic forecasting problem as a future frame prediction task with relatively weak temporal dependencies (might be due to stochastic urban traffic dynamics) and strong prior knowledge, \textit{i.e.}, the roadmaps of the cities.
For these reasons, we use the U-Net as the backbone model, and we propose a roadmap generation method to make the predicted traffic flows more rational. 
Meanwhile, we use a fine-tuning strategy based on the validation set to prevent overfitting, which effectively improves the prediction results. 
At the end of this report, we further discuss several approaches that we have considered or could be explored in future work: (1) harnessing inherent data patterns, such as seasonality; (2) distilling and transferring common knowledge between different cities. We also analyze the validity of the evaluation metric. Code is available at \url{https://github.com/ZJianjin/Traffic4cast2020_LDS}.
\end{abstract}

\section{Introduction}
Traffic forecasting is an important problem in some practical applications, such as the calculation of the \textit{estimated time of arrival} (ETA) and an autonomous driving system \cite{DBLP:conf/cvpr/GuptaJFSA18,DBLP:conf/cvpr/AlahiGRRLS16,DBLP:conf/iccv/IvanovicP19}, which is a hot-spot topic in the current computer vision research.
It is also challenging due to the high dimensional nature of the spatial size of data and the complexity and randomness of urban traffic changes.
Following the mainstream deep learning models for deterministic future frames prediction~\cite{DBLP:conf/icml/SrivastavaMS15,DBLP:conf/nips/DentonB17,DBLP:conf/nips/FinnGL16}, we take the past traffic flows as input frames and view the forecasting problem as a unique prediction task of future frames with some domain-specific prior knowledge.


Inspired by the first place solution~\cite{DBLP:journals/corr/abs-1912-05288} to the challenge last year, we develop our model on the base of U-Net~\cite{DBLP:conf/miccai/RonnebergerFB15}.
The major advantage is that compared to the LSTM-based recurrent prediction model, U-Net is more efficient in memory footprint and training time.
Considering that the evaluation metric is the \textit{mean squared error} (MSE), we design the U-Net to be a deterministic prediction model.
The GAN-based or VAE-based training schemes~\cite{DBLP:conf/cvpr/Tulyakov0YK18,denton2018stochastic} are not used, which might potentially affect the prediction accuracy in terms of MSE.

In this work, we explore solutions of transfer learning to mitigate the gap between the training and test sets. 
Specifically, we propose a fine-tuning strategy that fully utilizes the validation data for traffic forecasting.
Furthermore, we consider the road structure as prior knowledge and refine the generated frames with roadmap masks. This approach is model-agnostic and is shown to effectively improve the prediction accuracy.
We present corresponding ablation studies to demonstrate the improvements brought by the above-mentioned tricks.
In this technical report, we also present an analysis of whether the current evaluation metric is appropriate or not. 
Meanwhile, we discuss how to fully utilize the inherent data pattern to facilitate the prediction task, which are promising directions for future work.

\section{Data Preprocessing} 
The traffic data is collected from $3$ cities (Berlin, Moscow, and Istanbul).
Both inputs and outputs are in the format of video frames. 
Specifically, we denote the shape of inputs and outputs as $T_{in}\times W\times H \times C_{in}$ and $T_{out}\times W\times H \times C_{out}$ respectively.
We have $T_{in}=12, T_{out}=6, W=495, H=436, C_{in}=9, C_{out}=8$.

For each pixel in the video frame, the value encodes the traffic information, i.e., speed and volume. 
The value of each pixel corresponds to the $5$-minute traffic information in an area of $100$m$\times100$m.
The pixel in the input frame has $9$ channels, which consists of the traffic speed and volume information in $4$ different directions (northwest, northeast, southwest, and southeast).
The $9$-th channel represents the level of traffic events.
The outputs are the first $8$ channels.
Different from the previous year, the outputs correspond to timestamps of $5$, $10$, $15$, $30$, $45$, $60$ minutes.

\begin{figure}[h]
  \centering
  \includegraphics[width=\columnwidth]{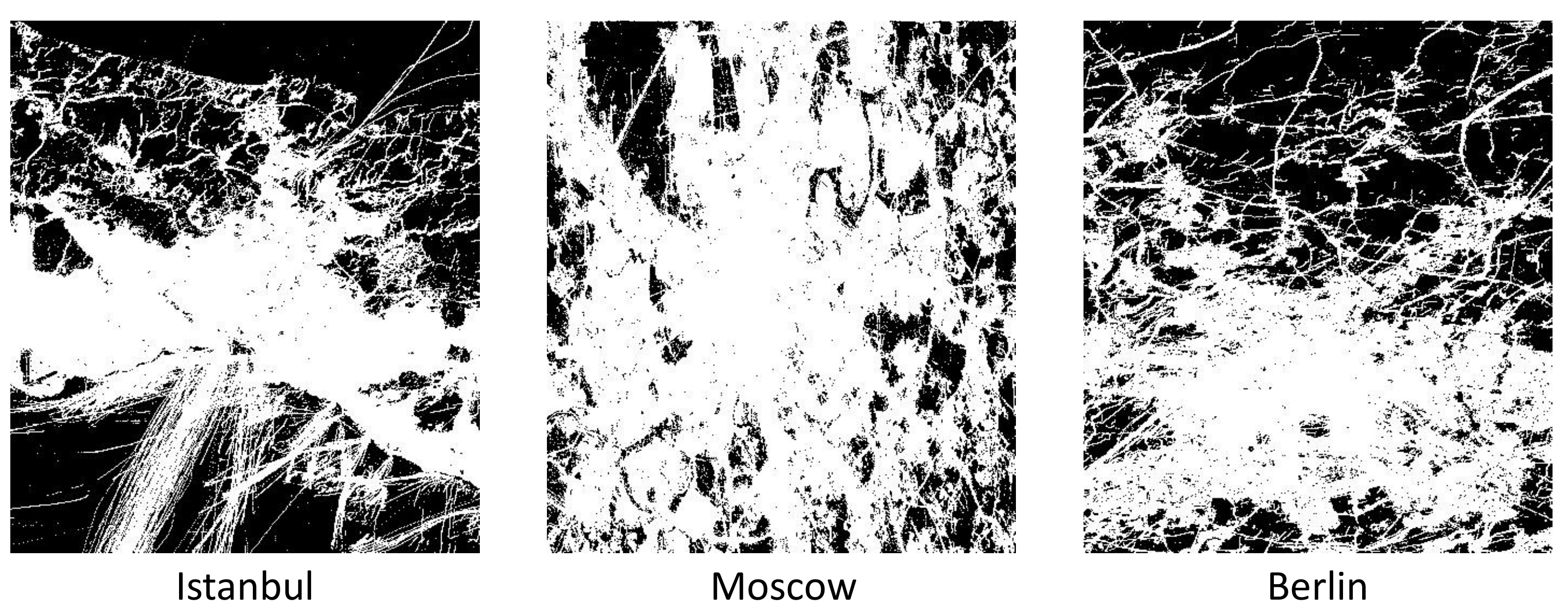}
  \caption{The obtained road mask on three cities. From left to right: Istanbul, Moscow, and Berlin.}
  \label{fig:Mask}
\end{figure}

\paragraph{Roadmap mask generation.}
As a common prior knowledge, we can only drive on the roads. However, the traffic information predicted by the U-Net~\cite{DBLP:conf/miccai/RonnebergerFB15} is unconstrained, which might be out of the road. To this end, we generate a road mask frame for each city and add them to the results. Specifically, for each city and each direction, we calculate the average traffic speed over the whole training set. If the value of a pixel in the average map is larger than $0$, this position is set as $1$. We thus have $4$ maps ($4$ directions) for each city. Here we present the obtained road masks as shown in Figure~\ref{fig:Mask}. The mask is directly multiplied to the outputs, which turned to be effective in improving the accuracy of predicted results.

\section{Approach}
\begin{figure}[h]
  \centering
  \includegraphics[width=\columnwidth]{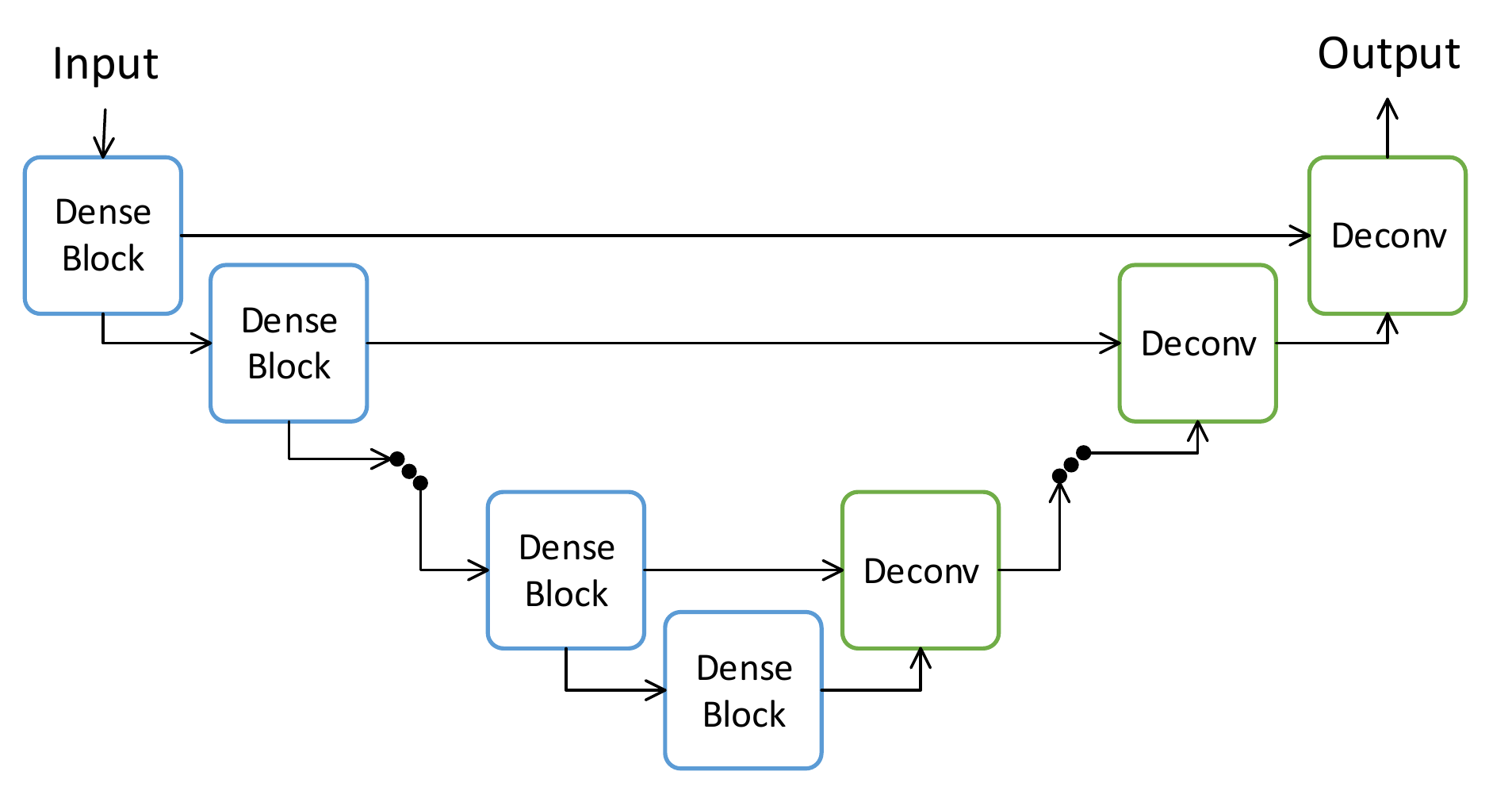}
  \caption{The model architecture of U-Net~\cite{DBLP:conf/miccai/RonnebergerFB15}. It have $8$ dense blocks and $7$ transpose convolutional layers. 
  The output of the $n$-th dense block is the input of the $(7-n)$-th transpose convolutional layer.}
  \label{fig:unet}
\end{figure}

\paragraph{Network architecture. }
We implement our model based on U-Net~\cite{DBLP:conf/miccai/RonnebergerFB15}, which is also applied in the Traffic4cast 2019 competition and achieved the best results. As shown in Figure~\ref{fig:unet}, it consists of $8$ dense blocks and $7$ transpose convolutional layers. The output of the $n$-th dense block will be input to the $(7-n)$-th transpose convolutional layer, which is a shortcut to retrieval more information from the original image. We also use these shortcuts inside the dense blocks. Each dense block has $4$ convolutional layers. Each layer will take the output of the previous layer and the original input of the dense block as inputs, following~\cite{DBLP:conf/cvpr/HuangLMW17}. Dense blocks are followed by average pooling to half the size of the feature maps. The last dense block is followed by a single convolutional layer.

\paragraph{Two-stage training strategy.} 
We find that the training data expands over $163$ days of the whole year, which is from January to June. However, as indicated by the time slot information, our model is tested in the second half of the year. The pattern of traffic flow might be very different during different seasons. We thus try to use the seasonal information to facilitate the prediction, which, however, got no obvious gain. We will give more details in Section~\ref{sec:dis}. We notice the validation set is from the second half of the year. So we designed a two-stage training strategy. Specifically, after pre-trained on the training set, the model will be fine-tuned on validation data because it is closer to testing data than training data. 

\section{Experiment}

\paragraph{Training details.} We choose the mean square error as the loss function following the evaluation metric and use the Adam optimizer with a $3e^{-4}$ learning rate.  For each city, we have $181$ train data files which contain traffic data of half a year, and $18$ files for validation. As mentioned above, each file has $288$ frames with a size of $495\times436\times9$. We take $12$ consequent frames as the inputs and predict $6$ frames as the outputs. In addition to the original frames, we also take the static data, which represents road intersections, as additional input features. All the input frames are normalized to $0$--$1$.

For each city, we train an independent model. As we described above, each model has a two-stage training process. All of them are trained for $5$ epochs and then fine-tuned for $1$ epoch.

\paragraph{Results.} Shown as Table \ref{tab:final_result}, the bottom row is the final model. As an ablation study, we remove the fine-tuned process (the second row) or road mask (the third row) independently for comparison. We can see that the fine-tuning process has a $0.000011$ gain and the road mask have a $0.000002$ gain.
Compared to the baseline model (the first row), the roadmap mask (the second row) effectively improves the prediction accuracy by filtering out the impossible driving area.
Meanwhile, the proposed two-stage training scheme reduces the gap between the training and testing sets, which boosts the prediction performance by a large margin (the third row).
Combing the above two aspects (the bottom row) achieves better results for the traffic forecasting task.

\begin{table}[h]
    \begin{center}
    \caption{Results on the test set}
    \label{tab:final_result}
    \begin{tabular}{lc}
    \toprule
    Model & MSE\\
    \midrule
    U-Net~\cite{DBLP:conf/miccai/RonnebergerFB15}  & 0.00119438  \\
    U-Net~\cite{DBLP:conf/miccai/RonnebergerFB15} + Roadmap mask & 0.00117991 \\
    U-Net~\cite{DBLP:conf/miccai/RonnebergerFB15} + Two-stage training & 0.00117037 \\
    Final model & 0.00116868 \\ 
    \bottomrule
    \end{tabular}
    \end{center}
\end{table}

\section{Discussion and Future Work} 
\label{sec:dis}
We discuss the future directions of further improving our work in terms of three aspects, i.e., the inherent pattern within the traffic data, the transferability of models learned with different data, and the evaluation metric.

\paragraph{Seasonality. }
Traffic changes are implicitly related to the season. 
For example, Moscow has such a long-term winter that the traffic speed tends to be low when it is snowing. 
Such a kind of seasonality contained in the traffic data could be utilized to facilitate traffic forecasting.
We train a model with the traffic data (Moscow) in January, February, and March. 
We then test on the data in November and December.
The predicted results demonstrate the effectiveness of the seasonal pattern.
However, for the other two cities (Istanbul and Berlin), where the seasonal data is rare and hard to capture, the prediction results are not so satisfying.
As the future work, we think such a direction is promising, i.e., utilizing the seasonal pattern for traffic forecasting, if sufficient data is available.

\paragraph{Knowledge transfer between cities.} 
In this technique report, we also have explored different methods to make full use of all the cities' data. Firstly, we try to fine-tune the target city using the well-pretrained source model directly. However, it could not perform well in the target city. Then we decide to adapt the transferable memory unit (TMU)~\cite{yao2020unsupervised} into the existing U-Net~\cite{DBLP:conf/miccai/RonnebergerFB15} model. The TMU is specifically designed for the unsupervised transfer learning problem using multiple pre-trained models to improve the performance of a new spatiotemporal predictive learning task. It could distill useful knowledge from different source models. While using TMU can perform better than directly fine-tuning, it is still hard to achieve the expected performance for two reasons: (1) Our models are CNN-based while the TMU is used for RNN-based models. (2) Different cities may have unique road network information. How to grasp the information of road networks for each city is crucial for traffic forecasting. While it is still difficult to capture the road network's domain-specific knowledge of each city, capturing the shared knowledge between cities seems much harder. As the future work, we think this direction is still promising. We may explore more effective ways to transfer knowledge from source cities.

\begin{figure}[h]
  \centering
  \includegraphics[width=\columnwidth]{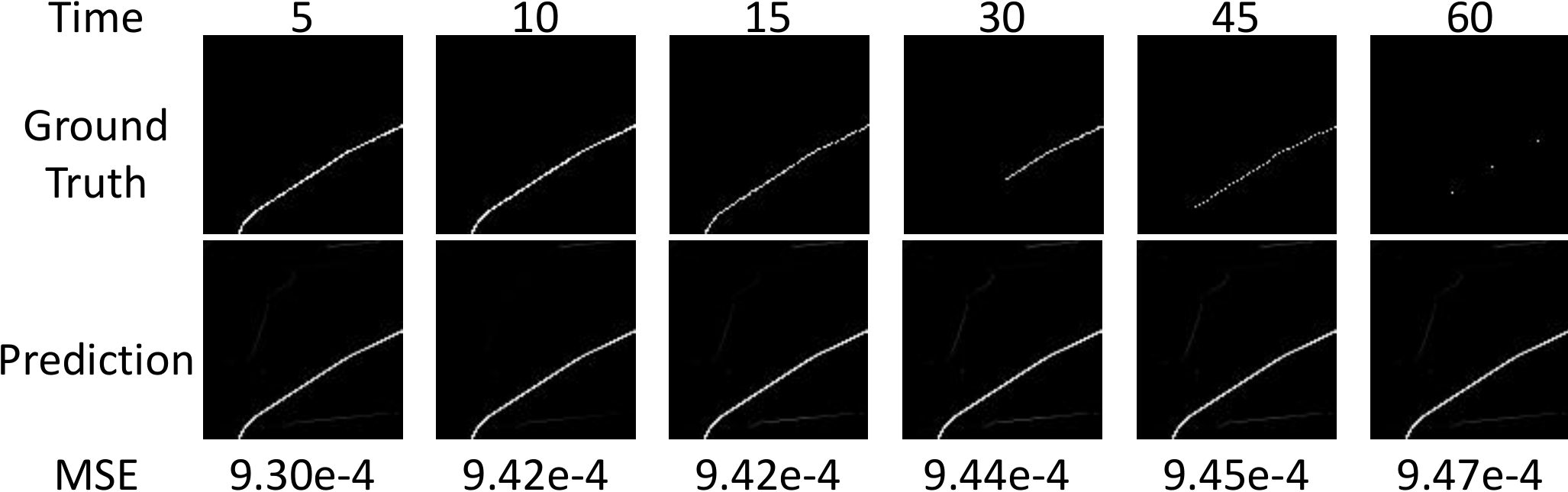}
  \caption{The predicted results at timestamp $5$, $10$, $15$, $30$, $45$, $60$ minutes. The predicted traffic behaviour is largely different from the ground truth but with low mean squared error.}
  \label{fig:vis}
\end{figure}

\textbf{Evaluation metric.} As a long-standing problem in the video prediction task, the evaluation metric based on mean squared error (MSE) might not fully reflect the performance of the prediction model.
As shown in Figure~\ref{fig:vis}, the first row is the ground truth sequence, while the second row refers to the predicted sequence.
This sequence is selected from the top 10$\%$ of the validation sequence in terms of MSE.
However, we can see that the predicted results are visually quite different from the ground truth.
As future work, it requires more attention and effort to find more appropriate evaluation metrics for this task.

\bibliographystyle{unsrt}
\bibliography{cites.bib}

\end{document}